\documentclass[sigconf,nonacm]{acmart}

\usepackage{subfigure}
\usepackage{tabularx}
\usepackage{diagbox}

\AtBeginDocument{%
  \providecommand\BibTeX{{%
    \normalfont B\kern-0.5em{\scshape i\kern-0.25em b}\kern-0.8em\TeX}}}

\setcopyright{none}
\renewcommand\footnotetextcopyrightpermission[1]{}
\setcopyright{none}
\makeatletter
\renewcommand\@formatdoi[1]{\ignorespaces}
\makeatother

\begin{document}

\title{Efficient Graph Deep Learning in TensorFlow with tf\_geometric}

\author{Jun Hu$^{1}$, Shengsheng Qian$^{1,2}$, Quan Fang$^{1,2}$, Youze Wang$^{3}$}
\author{Quan Zhao$^{3}$, Huaiwen Zhang$^{1,2,4}$, Changsheng Xu$^{1,2}$}
\affiliation{
  \institution{$^{1}$National Laboratory of Pattern Recognition, Institute of Automation, Chinese Academy of Sciences}
  \country{}
}
\affiliation{
  \institution{$^{2}$School of Artificial Intelligence, University of Chinese Academy of Sciences}
  \country{}
}
\affiliation{
  \institution{$^{3}$Hefei University of Technology}
  \country{}
}
\affiliation{
  \institution{$^{4}$Key Laboratory of Artificial Intelligence Scenario Application and Intelligent System Evaluation (China Center for Information Industry Development), Ministry of Industry and Information Technology}
  \country{}
}
\email{hujunxianligong@gmail.com, {shengsheng.qian, qfang}@nlpr.ia.ac.cn}
\email{{youze.wang,2019111002}@mail.hfut.edu.cn, {huaiwen.zhang,csxu}@nlpr.ia.ac.cn}

\renewcommand{\shortauthors}{Jun Hu, Shengsheng Qian, Quan Fang, Youze Wang, Quan Zhao, Huaiwen Zhang, Changsheng Xu}


\begin{abstract}
We introduce tf\_geometric\footnote{https://github.com/CrawlScript/tf\_geometric}, an efficient and friendly library for graph deep learning, which is compatible with both TensorFlow 1.x and 2.x.
tf\_geometric provides kernel libraries for building Graph Neural Networks (GNNs) as well as implementations of popular GNNs.
The kernel libraries consist of infrastructures for building efficient GNNs, including graph data structures, graph map-reduce framework, graph mini-batch strategy, etc.
These infrastructures enable tf\_geometric to support single-graph computation, multi-graph computation, graph mini-batch, distributed training, etc.; therefore, tf\_geometric can be used for a variety of graph deep learning tasks, such as transductive node classification, inductive node classification, link prediction, and graph classification.
Based on the kernel libraries, tf\_geometric implements a variety of popular GNN models for different tasks.
To facilitate the implementation of GNNs, tf\_geometric also provides some other libraries for dataset management, graph sampling, etc.
Different from existing popular GNN libraries, tf\_geometric provides not only Object-Oriented Programming (OOP) APIs, but also Functional APIs, which enable tf\_geometric to handle advanced graph deep learning tasks such as graph meta-learning.
The APIs of tf\_geometric are friendly, and they are suitable for both beginners and experts.
In this paper, we first present an overview of tf\_geometric's framework.
Then, we conduct experiments on some benchmark datasets and report the performance of several popular GNN models implemented by tf\_geometric.

\end{abstract}




\maketitle

\section{Introduction}

Graph is a powerful data structure that can be used to model relational data, and it is widely used by real-world applications.
In recent years, Graph Neural Networks (GNNs) emerge as powerful tools for deep learning on graphs, which aims to understand the semantics of graph data.
GNNs have been successfully applied to a variety of tasks in different fields, such as recommendation systems~\cite{DBLP:conf/www/Fan0LHZTY19,DBLP:conf/www/WangZLLZLZ20,DBLP:conf/www/TanLZYZH20}, question answering systems~\cite{DBLP:conf/iccv/LiGCL19,DBLP:conf/emnlp/FangSGPWL20,DBLP:conf/mm/HuQFX19}, neural machine translation~\cite{DBLP:conf/emnlp/BastingsTAMS17,DBLP:conf/naacl/MarcheggianiBT18}, traffic prediction~\cite{DBLP:conf/www/Wang0WJWTJY20,DBLP:journals/tits/CuiHKW20}, drug discovery and design~\cite{DBLP:conf/nips/FoutBSB17,DBLP:journals/corr/abs-2012-05716}, diagnosis prediction~\cite{DBLP:journals/bigdata/LiQZL20,DBLP:journals/corr/abs-2101-02870}, and physical simulation~\cite{pfaff2021learning,DBLP:conf/icml/Sanchez-Gonzalez20}.

\begin{figure}[t]
  \centering\includegraphics[width=3.0in]{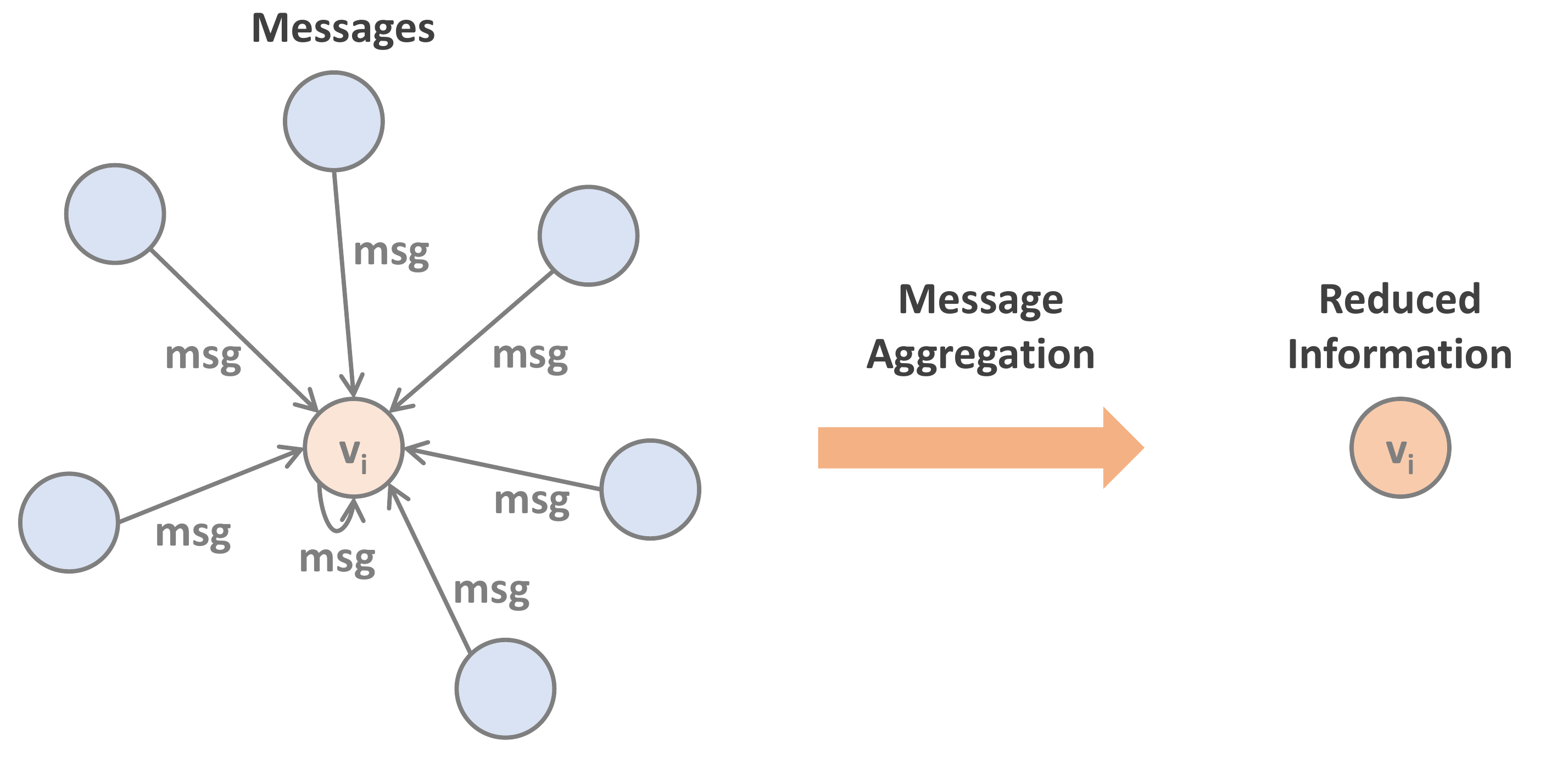}
  \vspace{-2mm}
  \caption{An Example of Message Aggregation}
  \vspace{-4mm}
   \label{fig:msg_aggr}
  \end{figure}

\begin{figure*}[t]
  \centering
  \subfigure[Common Graph Pooling]{
    \label{fig:common_graph_pooling} 
    \includegraphics[width=2.1in]{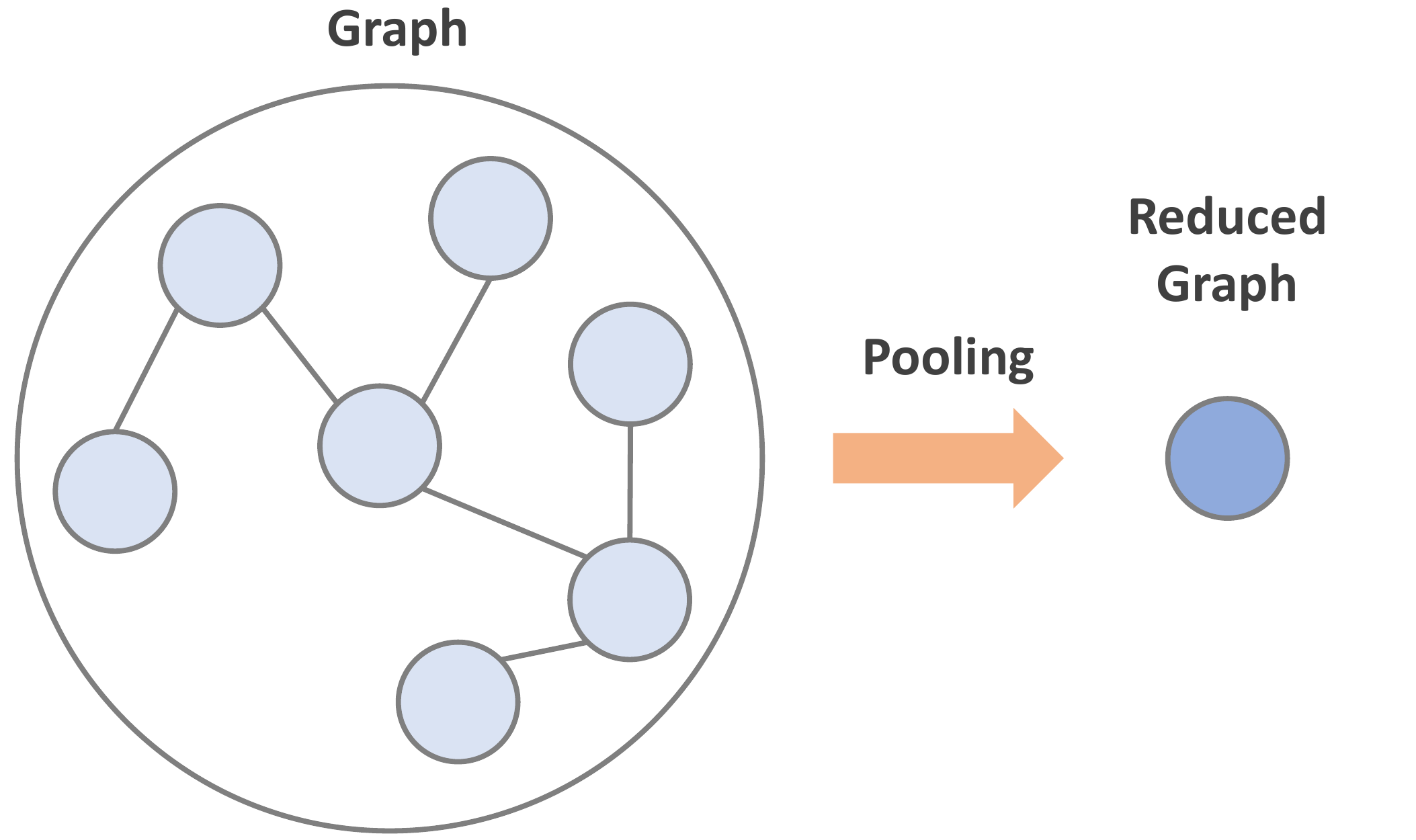}
    }
\hspace{0.5in}
  \subfigure[Hierarchical Graph Pooling]{
    \label{fig:hierarchical_graph_pooling} 
    \includegraphics[width=4.0in]{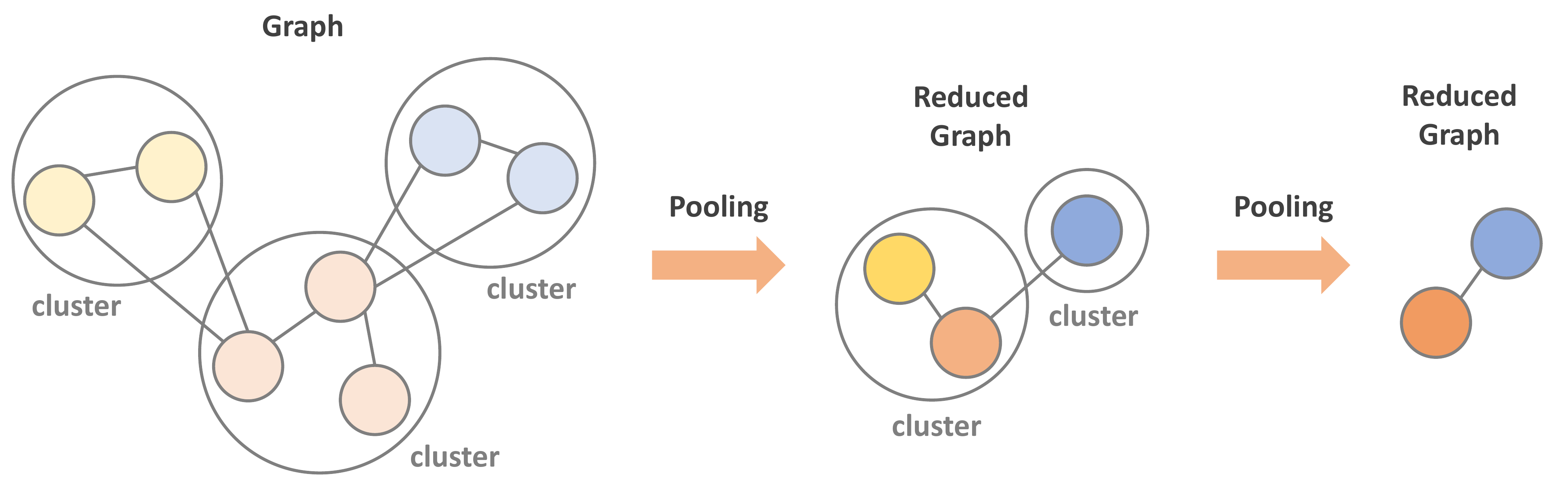}
    }

  \vspace{-1mm}
\caption{Examples of Graph Pooling}
\vspace{-1mm}
  \label{fig:graph_pooling} 
\end{figure*}

Due to the properties of graph data, such as sparsity and irregularity, it is challenging to implement efficient and friendly GNN libraries.
It is known that the most challenging problem for implementing GNNs is the aggregation of graph data.
%
%
Aggregation is the fundamental operation for most GNNs.
There are mainly two types of aggregation in GNNs: message aggregation and graph pooling.
(1) The \textbf{message aggregation}, which is also called message passing, aims to aggregate multiple messages between a node and its context and reduce them into one element.
Fig.~\ref{fig:msg_aggr} shows an example of message aggregation in Graph Convolutional Networks (GCNs).
In the example, the context of node $v_i$ consists of its neighbor nodes and itself.
The context nodes pass multiple messages to $v_i$ and these messages are reduced to a feature vector, which is then used as the high-order representation of node $v_i$.
(2) The \textbf{graph pooling} aims to aggregate elements in graphs or clusters and reduce them into high-order graph-level or cluster-level representations.
Fig.~\ref{fig:common_graph_pooling} shows an example of graph pooling for learning graph-level representations.
The representations of all the nodes in the graph are aggregated to generate the representation of the graph.
In some complex graph pooling models, the graph pooling layers are used to obtain a pooled graph rather than a graph representation vector~\cite{DBLP:conf/nips/YingY0RHL18,DBLP:conf/icml/LeeLK19,DBLP:conf/aaai/RanjanST20}.
For example, as shown in the hierarchical graph pooling example in Fig.~\ref{fig:hierarchical_graph_pooling}, the graph pooling operation aggregates nodes in each cluster and reduces them as a node in the pooled graph.
The two main types of aggregation allow researchers and engineers to design complex GNNs, and thus proper solutions for aggregation on graphs are imperative for building elegant GNN models.
However, it is non-trivial to design proper aggregation solutions for sparse and irregular graph data.
Most intuitive solutions, such as padding and masking, usually suffer from the memory and efficiency problem, whereas many efficient solutions, such as sparse matrix multiplication, require a lot of complex tricks to accomplish advanced aggregations.
Moreover, most efficient solutions require users to use specific data structures to organize the graph data.

We develop tf\_geometric, an efficient and friendly GNN library for deep learning on sparse and irregular graph data, which is compatible with both TensorFlow 1.x and 2.x.
tf\_geometric provides a unified solution for GNNs, which mainly consists of kernel libraries for building graph neural networks and implementations of popular GNNs.
The kernel libraries contain infrastructures for building efficient GNNs, including graph data structures, graph map-reduce framework, graph mini-batch strategy, etc.
In particular, the graph data structure and map-reduce framework provide an elegant and efficient way for aggregation on graphs.
The kernel libraries enable tf\_geometric to support single-graph computation, multi-graph computation, graph mini-batch, distributed training, etc.; therefore, tf\_geometric can be used for a variety of graph deep learning tasks, such as transductive node classification, inductive node classification, link prediction, and graph classification.
Based on the kernel libraries, a variety of popular GNN models for different tasks are implemented as APIs for tf\_geometric.
To facilitate the implementation of GNNs, tf\_geometric also provides some other libraries for dataset management, graph sampling, etc.
Different from existing popular GNN libraries, tf\_geometric provides not only OOP APIs, but also Functional APIs, which enable tf\_geometric to handle advanced graph deep learning tasks such as graph meta-learning.
The APIs of tf\_geometric are friendly, and they are suitable for both beginners and experts.
tf\_geometric is available on GitHub\footnote{https://github.com/CrawlScript/tf\_geometric}.
The features of tf\_geometric are thoroughly documented\footnote{https://tf-geometric.readthedocs.io} and a collection of accompanying tutorials and examples are also provided in the documentation.

\section{Overview}

tf\_geometric mainly consists of kernel libraries for building graph neural networks and implementations of popular GNNs.
Besides, some other libraries such as dataset management and graph sampling are also provided to facilitate the implementation of GNNs.
In this section, we provide an overview of different parts of the tf\_geometric.

\begin{figure}[b]
  \centering\includegraphics[width=3.3in]{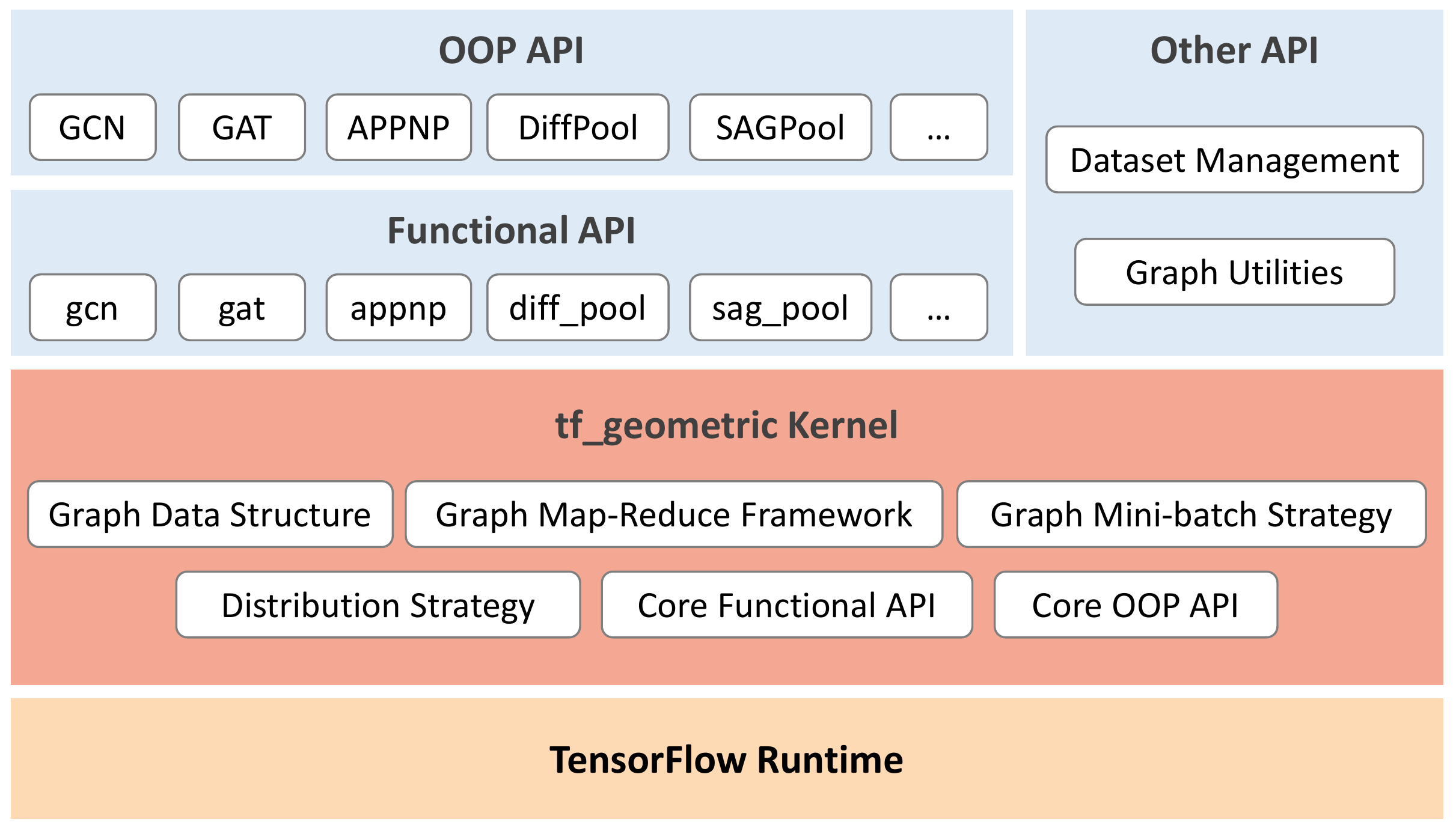}
  \caption{The Framework of tf\_geometric}
   \label{fig:framework}
  \end{figure}

\begin{figure*}[t]
  \centering\includegraphics[width=7.0in]{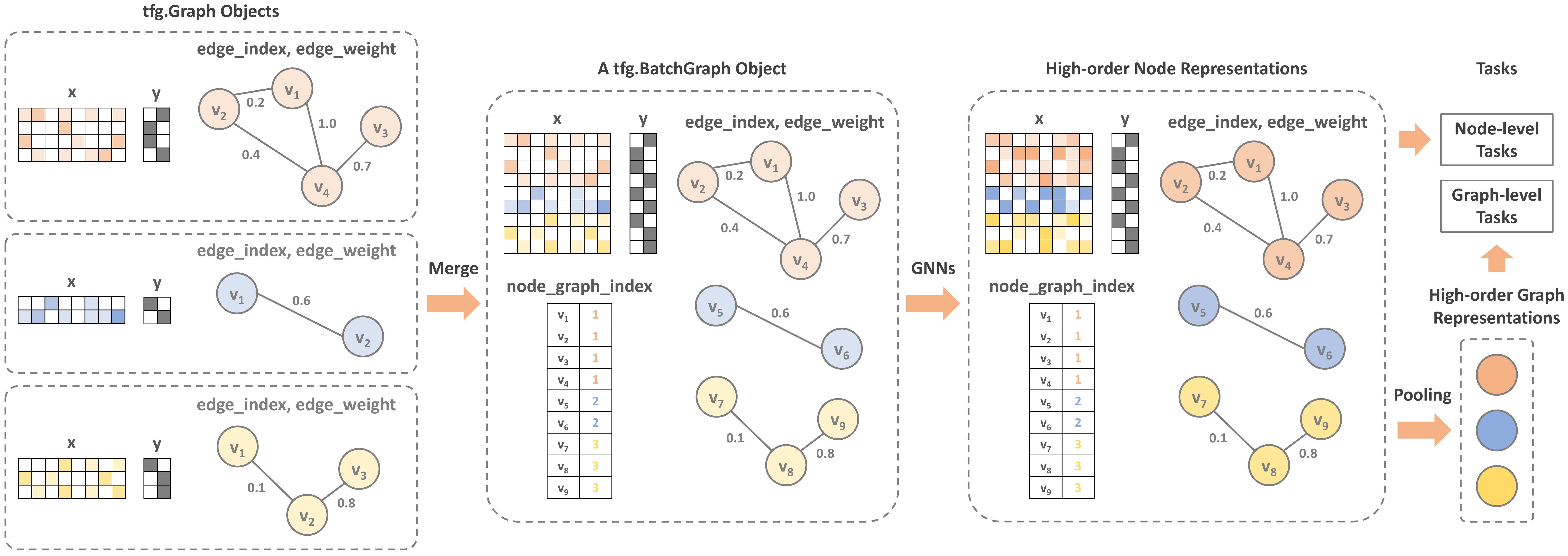}
  \vspace{-7mm}
  \caption{tfg.Graph and tfg.BatchGraph}
  \vspace{-3mm}
   \label{fig:graph_and_batch_graph}
  \end{figure*}

\subsection{Kernel Libraries}

The framework of the kernel libraries is shown in Fig.~\ref{fig:framework}.
As shown in Fig.~\ref{fig:framework}, the kernel libraries consist of several fundamental components as infrastructures for building efficient GNNs, including graph data structures, graph map-reduce framework, graph mini-batch strategy, etc.
These infrastructures enable tf\_geometric to support single-graph computation, multi-graph computation, graph mini-batch, distributed training, etc., and therefore tf\_geometric can be used for a variety of graph deep learning tasks, such as transductive node classification, inductive node classification, link prediction, and graph classification.
In this section, we will introduce these infrastructures in detail.

\subsubsection{Graph Data Structure}

tf\_geometric has two core graph data structures (classes): \textbf{Graph} and \textbf{BatchGraph}, which are used to model a single graph and a batch of graphs, respectively.
In this section, we first introduce some notations for graph data in graph deep learning and then show how tf\_geometric organizes graph data with its graph data structures.

Generally, a graph can be represented as $G=(V, E)$, where $V=\{v_1, v_2, ..., v_{|V|}\}$ is the set of nodes and $E=\{e_1, e_2, ..., e_{|E|}\}$ denotes the set of edges.
In graph deep learning, the graph $G$ is usually presented as $\mathcal{G}=(X, A)$, where $X$ and $A$ are the node feature matrix and adjacency matrix, respectively.
The node feature matrix $X \in \mathbb{R}^{|V| \times d}$ contains features of all the nodes in the graph, and its $i_{th}$ row represents the $d$-dimensional feature vector of the $i_{th}$ node in the graph.
The adjacency matrix $A \in \mathbb{R}^{|V| \times |V|}$ contains the edge information, where a positive entry $A_{ij}$ indicates these exists an edge from the $i_{th}$ node $v_i$ to the $j_{th}$ node $v_j$ with weight $A_{ij}$.
In some tasks, such as node classification and graph classification, the node label or graph label information is also required.
The label information is denoted as $Y$, and the graph can be further represented as $\mathcal{G}=(X, A, Y)$.
Usually, $Y$ is presented as a list of integer label indices or a matrix of encoded label vectors.

The \textbf{Graph} class is used to model a single graph.
A graph $\mathcal{G}=(X, A, Y)$ can be modeled as a Graph object $graph = \allowbreak (x, edge\_index, \allowbreak edge\_weight, y)$
%
%
, where $x$, $(edge\_index, \allowbreak edge\_weight)$, and $y$ correspond to $X$, $A$ and $Y$, respectively.
The node feature matrix $X$ and label information $Y$ are modeled as dense tensors $x$ and $y$ respectively, while the adjacency matrix $A$ is presented as a sparse matrix in coordinate (COO) format, which consists of the indices of entries $edge\_index \in \mathbb{R}^{2 \times |E|}$ and the values of entries $edge\_weight \in \mathbb{R}^{|E|}$.
It is known that many aggregation operations in GNNs can benefit a lot from the COO format sparse adjacency matrix.
Moreover, the COO format data is friendly to many advanced aggregation operations of TensorFlow, such as the \textbf{tf.math.segment\_xxxx} operations, which are important for building efficient GNN models.
Note that either the input data or the intermediate output tensors can be used to construct graph objects.
Especially, since the construction of Graph objects does not involve any deep copy operations, it is a differentiable operation that can be applied to any intermediate output tensors that require gradients. 
For each \textbf{Graph} object, the GNN models in tf\_geometric can take advantage of parallelism capabilities of deep learning frameworks to efficiently process information in the graph.
However, due to the limitation of most deep learning frameworks, it is difficult to process information in different graph objects simultaneously.
Therefore, for tasks dealing with multiple graphs, such as inductive node classification and graph classification, tf\_geometric introduces the \textbf{BatchGraph} class, which allow the GNN models to process information in multiple graphs in parallel.

A \textbf{BatchGraph} object stores the information in multiple graph objects, and it enables the parallel processing of data from different graphs by virtualizing multiple graphs as a single graph.
The BatchGraph class is a subclass of the Graph class, and it can be denoted as $batch\_graph = \allowbreak (x, edge\_index, \allowbreak edge\_weight, y, node\_graph\_index)$, 
%
where $x$, $edge\_index$, $edge\_weight$, and $y$ are attributes inherited from the superclass Graph, and $node\_graph\_index$ is a list of integers indicating which graph each node belong to.
tf\_geometric first leverages a reindexing trick to reassign indices to nodes from different graphs such that each node has a unique index in the BatchGraph.
The left part of Fig.~\ref{fig:graph_and_batch_graph} shows an example of combining multiple Graph objects into a BatchGraph object.
The $j_{th}$ node of the $i_{th}$ graph is reindexed by adding an offset value, which is the number of nodes in the previous $i-1_{th}$ graphs.
In the example, the offset of the second graph and third graph is 4 and 6 respectively.
Therefore, the index of the first node in the second graph is reindexed as $1 + 4 = 5$, and the index of the second node in the third graph is reassigned as $2 + 6 = 8$.
After reindexing, the attributes of the graphs such as $x$ and $edge\_index$ are then adjusted and combined based on the reassigned node indices.
For example, the $edge\_index$ of the given graphs are first replaced with the reindexed node index and they are then stacked together to form a new $edge\_index$ for the BatchGraph.
Note that converting Graph objects into a BatchGraph object does not modify node features and the connectivity between nodes.
Thus, for most GNNs, applying them on multiple graphs iteratively is equivalent to applying them on the corresponding BatchGraph.
As a result, applying GNN operations on a BatchGraph automatically enables the parallel processing of multiple graphs, which brings dramatic performance improvement in computational efficiency.
As shown in the right part of Fig.~\ref{fig:graph_and_batch_graph}, the GNNs learn high-order features for nodes from different graphs, which can be further used for different multi-graph tasks.
For example, the learned node features can be directly used for inductive node classification tasks.
Moreover, the learned node features can be aggregated into graph representations (graph pooling) based on the $node\_graph\_index$ of the BatchGraph, which can be used for graph-level tasks such as graph classification.

\begin{figure}[t]
  \centering
  \subfigure[Map-Reduce Workflow for Normalized Attention Scores in GAT]{
    \label{fig:map_reduce} 
    \includegraphics[width=3.3in]{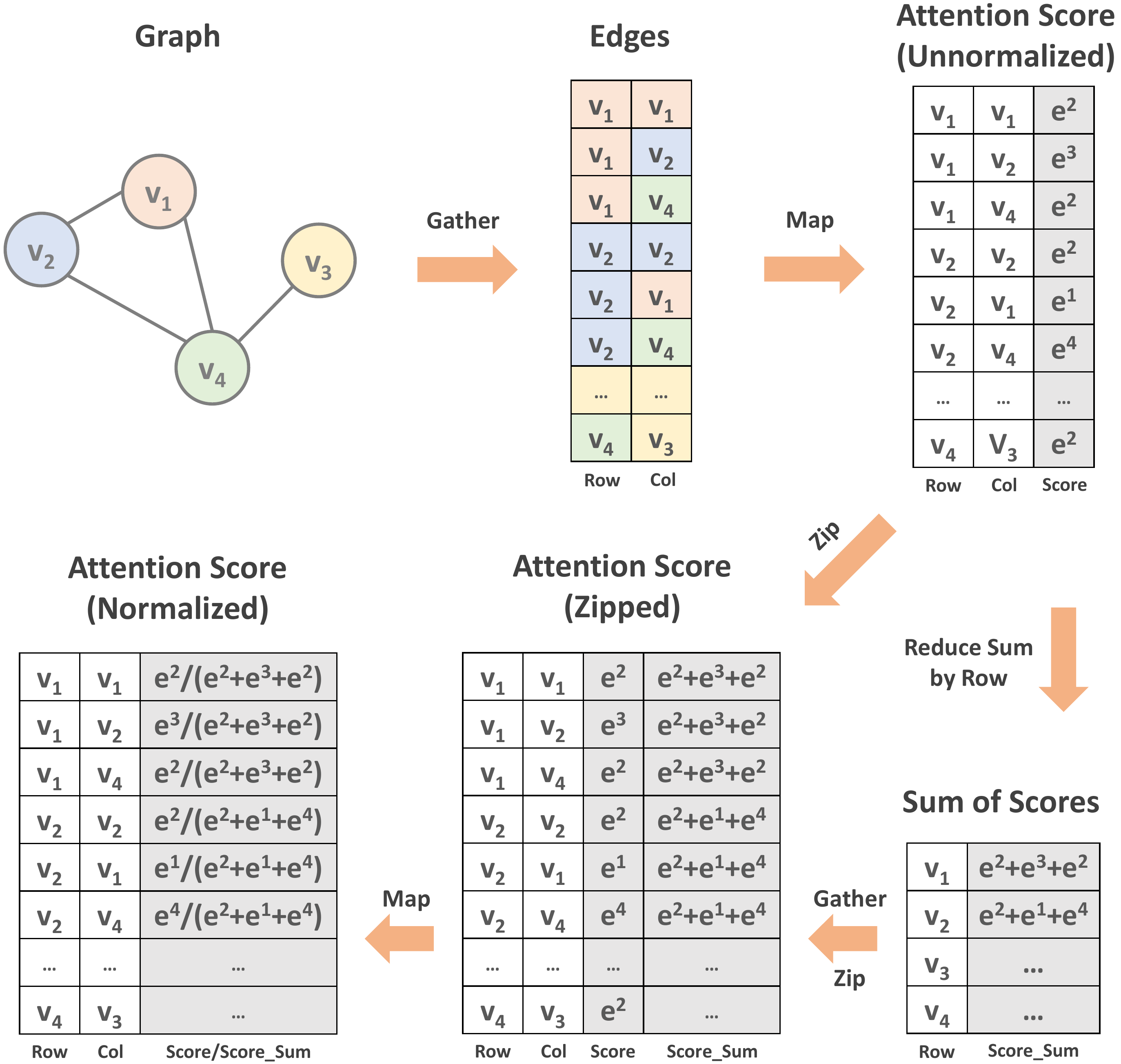}
    }
%
  \subfigure[Reduce by Key]{
    \label{fig:reducer} 
    \includegraphics[width=3.1in]{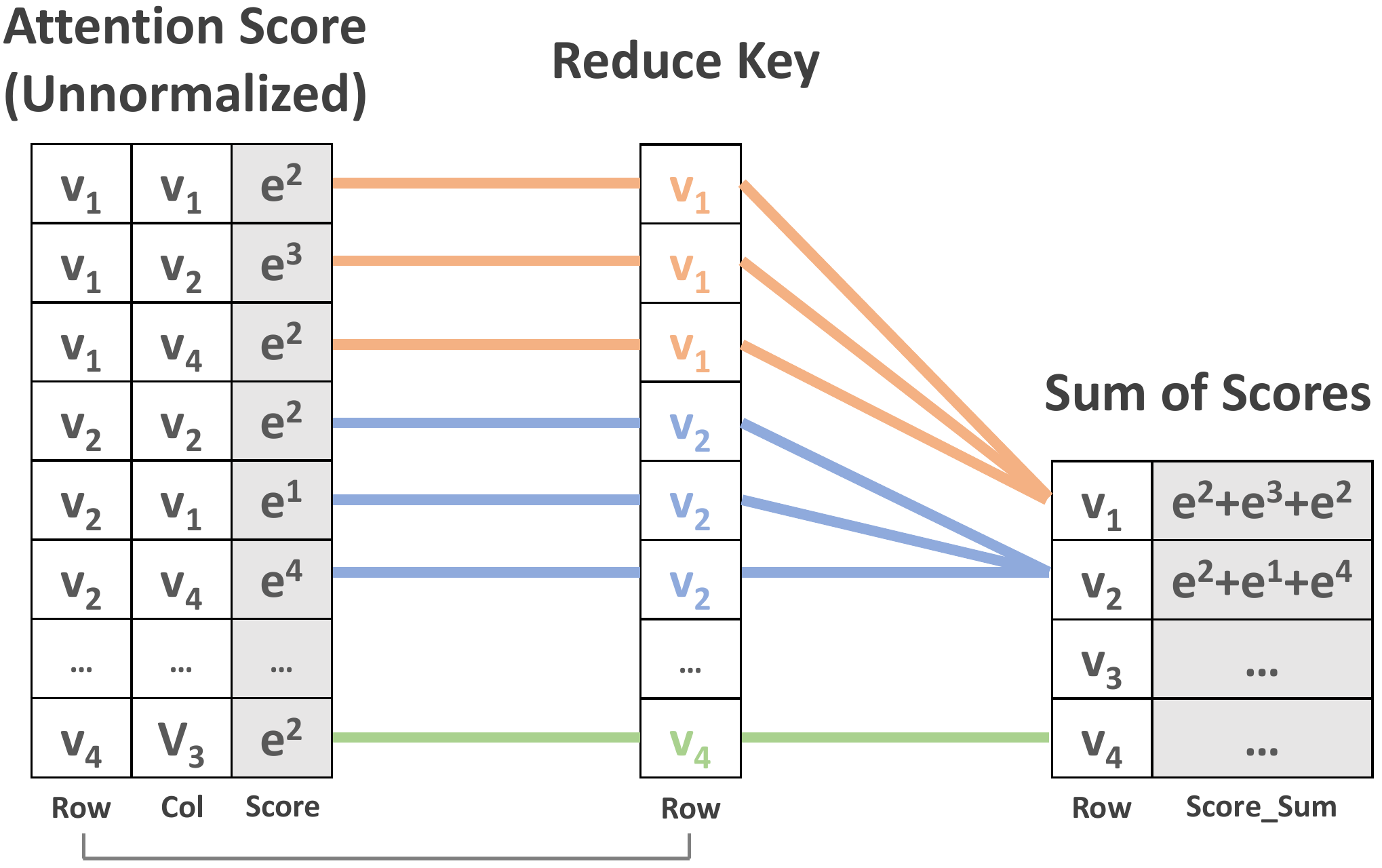}
    }

\vspace{-4mm}
\caption{Graph Map-Reduce Framework}
\vspace{-6mm}
\end{figure}

\subsubsection{Graph Map-Reduce Framework}

Many complex GNNs can be considered as a combination of simple map-reduce operations on graphs.
Usually, map and reduce operations correspond to transformation and aggregation operations on graphs.
The tf\_geometric kernel provides a graph map-reduce framework, including basic and advanced map and reduce operations on graphs and graph map-reduce workflows.

Fig.~\ref{fig:map_reduce} shows an example of map-reduce workflow, which computes the normalized attention scores for a Graph Attention Network (GAT).
In the example, each colored edge contains the feature vectors of a node and one of its neighbor nodes.
The first map operation parallelly transforms a batch of edges into unnormalized attention scores.
Map operations do not involve the interaction between elements, and most of them can be implemented with general TensorFlow operations.
Here, the first map operation is implemented with a TensorFlow dense layer.
GAT requires the attention score to be normalized by softmax normalization.
To achieve this, a reduce operation is introduced to aggregate the unnormalized attention scores of neighbors and obtain the denominator for the softmax normalization.
Details of the reducer are shown in Fig.~\ref{fig:reducer}.
The reducer aggregates information for each group and the reduce key indicates which group each element belongs to.
In this case, the reduce key is node index and the unnormalized attention scores of the neighbor nodes of a node share the same reduce key.
Different from common deep learning models for images and text, where most reduce operations are designed for tensors with regular shapes, the graph deep learning models usually require the reducer to deal with irregular data.
Thus, many general reduce operations, such as \textbf{tf.reduce\_sum} and \textbf{tf.reduce\_max}, can not be used as reducers for GNNs.
To address this problem, tf\_geometric takes advantage of several advanced APIs of TensorFlow to build efficient reducers for irregular graph data.
In the example, the sum reducer is implemented with the \textbf{tf.math.unsorted\_segment\_sum} API, and it can efficiently aggregate the unnormalized attention scores from different numbers of neighbors for each node.
The last map operation is easier than the aforementioned operations, and it can be implemented by a simple TensorFlow division function.

\subsubsection{Graph Mini-batch Strategy}

As with general deep learning models, GNNs can benefit from mini-batch training and inference of graphs.
Given a batch of graphs, tf\_geometric combine them into a BatchGraph and apply GNNs on it.
Since the label information $y$ is also combined in the BatchGraph, the combined $y$ can directly be used as the node/graph labels of the batch.

The mini-batch strategy in tf\_geometric is flexible, and you can mini-batch not only the input graph data, but also the intermediate output graphs.
In the mini-batch process, since the operation of combining graphs into a BatchGraph is differentiable, the gradients will pass from the BatchGraph back to the given batch of graphs.
Moreover, the mini-batch construction operation is fast enough and can be executed during each forward propagation process.

\subsubsection{Distribution Strategy}

To take advantage of the powerful distribution ability of TensorFlow, all the GNN models in tf\_geometric are implemented as standard TensorFlow models, which can be distributed with minimal code changes on the data processing.
The distribution of tf\_geometric GNN models can be easily handled by TensorFlow distribution strategies, and the model can be distributed in different ways by choosing different distribution strategies.
However, the distribution of graph data cannot be solved by simply applying the distribution strategies.
This is because the built-in data sharding mechanism of the distribution strategies are designed for regular tensors, which is not able to deal with irregular graph data.
Nonetheless, we can still easily distribute graph data by customizing distributed graph datasets with \textbf{tf.data.Dataset} for distribution strategies.
The customizing usually only requires a few small changes on the code for local data processing.

\subsubsection{Core OOP APIs and Functional APIs} tf\_geometric provides both \textbf{Object Oriented Programming (OOP) APIs} and \textbf{Functional APIs}, with which users can built advanced graph deep learning models:

\begin{itemize}

\item \textbf{OOP APIs} are class-level interfaces for graph neural networks.
The GNN classes in tf\_geometric are implemented as standard TensorFlow models by subclassing the \textbf{tf.keras.Model} class, where each GNN class defines how to maintain the model parameters and the computational process.
An instance of GNN classes holds the parameters of a GNN model and it can be called as a function to process the input data with the GNN algorithm.
OOP APIs are convenient since users can apply them on graph data as black boxes without knowing details of the GNN model, such as the initialization of model parameters and the algorithm.
Due to the convenience and customizability of OOP APIs, most popular GNN libraries provide OOP APIs as the main interface for GNNs.
However, OOP APIs are insufficient for some advanced tasks, and therefore tf\_geometric provides functional APIs to solve the problem, which will be introduced in the next paragraphs.

\item \textbf{Functional APIs} provide function-level interfaces for graph neural networks.
Functional APIs are functions that implement GNN operations.
Different from OOP APIs, which automatically maintain model parameters in GNN layer instances, functional APIs require users to maintain model parameters outside the GNN functions and use them as the input of GNN functions together with graph data.
That is, instead of using fixed tensors as model parameters in OOP APIs, functional APIs can dynamically use different tensors as model parameters for each call.
This feature of functional APIs is critical for advanced tasks that require complex maintenance strategies of model parameters, such as graph meta-learning.
For example, to implement MAML~\cite{DBLP:conf/icml/FinnAL17} on graphs, a GNN function will be called multiple times with different parameters during each forward propagation.
The GNN function is first called with variable tensors as initial parameters and then called multiple times with temporary tensors as updated model parameters.
Obviously, functional APIs are elegant solutions for this task, since dynamic parameters are natively supported by functional APIs.

\end{itemize}

Note that the core OOP APIs and functional APIs in the kernel do not involve the implementation of specific GNNs.
Instead, they provide some infrastructures that are essential for implementing specific GNN classes or functions, such as abstract classes and functions for graph map-reduce.

\vspace{-2mm}
\subsection{Implementation of Popular GNN Models}

Based on the kernel libraries, tf\_geometric implements a variety of popular GNN models for different tasks, including node-level models such as Graph Convolutional Network (GCN)~\cite{DBLP:conf/iclr/KipfW17}, Graph Attention Network (GAT)~\cite{DBLP:conf/iclr/VelickovicCCRLB18}, Simple Graph Convolution (SGC)~\cite{DBLP:conf/icml/WuSZFYW19}, Approximate Personalized Propagation of Neural Predictions (APPNP)~\cite{DBLP:conf/iclr/KlicperaBG19}, and Deep Graph Infomax (DGI)~\cite{DBLP:conf/iclr/VelickovicFHLBH19}, and graph-level models such as Set2Set~\cite{DBLP:journals/corr/VinyalsBK15}, SortPool~\cite{DBLP:conf/aaai/ZhangCNC18}, Differentiable Pooling (DiffPool)~\cite{DBLP:conf/nips/YingY0RHL18}, and Self-Attention Graph Pooling (SAGPool)~\cite{DBLP:conf/icml/LeeLK19}.
To avoid redundancy, all the GNN models in tf\_geometric are first implemented as Functional APIs, and the OOP APIs are just wrappers of the corresponding Functional APIs.
We carefully implement these models and make sure that they can achieve competitive performance with other implementations.

Besides, tf\_geometric also provides demos that reproduce the performance of GNNs reported in the literature.
The demos contain the complete code for data loading, training, and evaluation.
They are implemented in an elegant way and also act as the style guide for tf\_geometric.

\subsection{Dataset Management Mechanism}

tf\_geometric provides customizable dataset APIs and a lot of ready-to-use public benchmark datasets.

\subsubsection{Dataset Classes and Dataset Instances}

Each dataset has a corresponding dataset class and different instances of a dataset class (dataset instances) can represent different configurations for the same dataset.
Each dataset instance can automatically download the raw dataset from the Web and then pre-process it into convenient data formats, which can benefit not only tf\_geometric, but also other graph deep learning frameworks.
Besides, a caching mechanism is provided by dataset classes, which allow you to only process each raw dataset once and load it from the cache on-the-fly.

Dataset classes are not just simple wrappers of the raw graph datasets, and they may also involve complex feature engineering in the pre-processing.
For example, node degrees are frequently used features in graph classification tasks~\cite{DBLP:conf/kdd/WuHX19}.
By encapsulating the computation of node degrees in the pre-processing method of dataset classes, users can directly load node degrees as features from the datasets without considering the complex feature engineering process.

\subsubsection{Built-in Datasets}

The provided datasets, which are also called built-in datasets, consist of lots of public benchmark datasets that are frequently used in graph deep learning research.
Moreover, the built-in datasets cover datasets for various graph deep learning tasks, such as node classification and graph classification.

\subsubsection{Customizable Datasets}

Users can customize their datasets by simply subclassing built-in abstract dataset classes.
The built-in abstract dataset classes manage the workflow of dataset processing and already encapsulate the implementation of general processes, such as downloading, file management, and caching.
These general processes can be customized by the configuration parameters defined in subclasses, such as the URL of the dataset and whether the pre-processing result should be cached.
Since the data pre-processing processes are usually different across different datasets, the data pre-processing is defined as abstract methods in the superclasses and users can implement them for their datasets in the subclasses by overriding the abstract methods.

\subsection{Utilities}

Some important utilities are required for implementing graph deep learning models.
These utilities include tools for common graph data processing, type conversion, graph sampling, etc.
The tools are put in the utils module of tf\_geometric, and most of them are designed not only for tf\_geometric, but also for general graph deep learning implementations.

\section{Comparison to Other GNN Libraries}

In recent years, several GNN libraries have been developed for different deep learning frameworks.
Among them, popular libraries such as PyTorch Geometric (PyG)\footnote{https://github.com/rusty1s/pytorch\_geometric}~\cite{DBLP:journals/corr/abs-1903-02428} and Deep Graph Library (DGL)\footnote{https://github.com/dmlc/dgl}~\cite{DBLP:journals/corr/abs-1909-01315} have been widely used by researchers to deal with graph deep learning in different fields.
They provide extensible OOP APIs and implement a variety of GNN classes, with which users can easily handle general graph deep learning tasks.
As mentioned before, OOP APIs are insufficient for several advanced tasks such as graph meta-learning.
Different from these GNN libraries, tf\_geometric provide not only OOP APIs, but also Functional APIs, which can be used to deal with advanced graph deep learning tasks.
Moreover, popular GNN libraries for TensorFlow such as Spektral \footnote{https://github.com/danielegrattarola/spektral}~\cite{DBLP:journals/cim/GrattarolaA21} and StellarGraph \footnote{https://github.com/stellargraph/stellargraph} usually only support TensorFlow 2.x, whereas tf\_geometric is compatible with both TensorFlow 1.x and 2.x.
Furthermore, tf\_geometric provides a flexible and friendly caching system that can speed up some GNNs, while existing GNN libraries do not support caching or only support caching for few special cases.
For example, PyG only supports layer-level caching for GCN, which means that each PyG GCN layer with caching enabled is bound to a constant graph structure and usually it can only be used for transductive learning tasks on a single graph.  
Instead, the tf\_geometric GCN adopts a graph-level caching mechanism, and it can cache for different graph structures with different GCN normalization configurations.

\section{Empirical Evaluation}

To provide an overview of how GNN models implemented by tf\_geometric perform on common research scenarios, we conduct experiments with several public benchmark datasets on two different tasks.

\subsection{Tasks and Evaluation Metrics}

We evaluate several tf\_geometric models on two different tasks: node classification and graph classification.

\textbf{Node Classification}
We first conduct experiments on a semi-supervised node classification task with three benchmark datasets: Cora, CiteSeer, and Pubmed~\cite{DBLP:journals/aim/SenNBGGE08}.
We evaluate GCN~\cite{DBLP:conf/iclr/KipfW17}, GAT~\cite{DBLP:conf/iclr/VelickovicCCRLB18}, SGC~\cite{DBLP:conf/icml/WuSZFYW19}, APPNP~\cite{DBLP:conf/iclr/KlicperaBG19}, and DGI~\cite{DBLP:conf/iclr/VelickovicFHLBH19} on the task, where GCN, GAT, SGC, APPNP are end-to-end classification models, while DGI is a self-supervised model node representation learning model, where an extra logistic regression model is utilized for classification based on the learned node representations.
For the benchmark datasets, we use the same dataset splits as in~\cite{DBLP:conf/iclr/KipfW17}, where each dataset is split into a train set, a test set, and a validation set.
The validation set is used for early stopping and its label information is not used for training.
We report the classification accuracy scores on the test set.

\textbf{Graph Classification}
We also evaluate tf\_geometric on a graph classification task with three benchmark datasets: NCI1, NCI109~\cite{DBLP:journals/kais/WaleWK08}, and PROTEINS~\cite{Dobson2003DistinguishingES,Borgwardt2005ProteinFP}.
We evaluate several graph pooling models, including Mean-Max Pool, Set2Set~\cite{DBLP:journals/corr/VinyalsBK15}, SortPool~\cite{DBLP:conf/aaai/ZhangCNC18}, DiffPool~\cite{DBLP:conf/nips/YingY0RHL18}, and SAGPool$_h$~\cite{DBLP:conf/icml/LeeLK19}.
The Mean-Max Pool is a naive graph pooling model, which obtains graph representations by concatenating the mean pooling and max pooling results of GCNs.
These classification accuracy scores of these models are evaluated on three benchmark datasets using 10-fold cross-validation, where a training fold is randomly sampled as the validation set.
As with the node classification task, the validation set is only used for early stopping.
The architectures of graph pooling models are complex, and they may involve components other than the core graph pooling layers.
Some of these components are model-agnostic and can be utilized by some other GNN models to obtain better performance.
For example, the hierarchical graph pooling models may benefit from the mean-max pooling on both hidden and output layers, whereas the official implements may only consider using mean pooling.
Therefore, we update the architectures for some models for a fair comparison.

\begin{table}[t]
\centering
\caption{Performance on Node Classification Tasks.}
\vspace{-2mm}
\scalebox{0.90}{
\begin{tabular}{|l|c|c|c|}\hline
\diagbox{Model}{Dataset}  & Cora            & CiteSeer        & Pubmed          \\\hline
GCN                       & $81.7 \pm 0.5 $ & $71.0 \pm 0.8 $ & $79.1 \pm 0.6 $ \\\hline
GAT                       & $83.0 \pm 0.8 $ & $72.9 \pm 0.8 $ & $79.0 \pm 0.2 $ \\\hline
SGC                       & $81.1 \pm 0.0 $ & $72.1 \pm 0.0 $ & $79.2 \pm 0.1 $ \\\hline
APPNP                     & $83.6 \pm 0.7 $ & $72.2 \pm 0.9 $ & $79.0 \pm 0.7 $ \\\hline
DGI                       & $82.4 \pm 0.5 $ & $72.2 \pm 0.5 $ & $78.0 \pm 0.6 $ \\\hline
\end{tabular}
}
\vspace{-1mm}
\label{tab:node_classification}
\end{table}

\begin{table}[t]
  \centering
  \caption{Performance on Graph Classification Tasks.}
  \vspace{-2mm}
\scalebox{0.90}{
\begin{tabular}{|l|c|c|c|}\hline

\diagbox{Model}{Dataset}  & NCI1            & NCI109           &  PROTEINS        \\\hline
Mean-Max Pool           & $76.03 \pm 0.7$ & $75.62 \pm 0.7$  &  $74.18 \pm 0.6$ \\\hline
Set2Set                 & $72.01 \pm 0.6$ & $69.54 \pm 0.6$  &  $71.69 \pm 0.6$ \\\hline
SortPool                & $74.65 \pm 0.6$ & $73.22 \pm 0.7$  &  $68.09 \pm 2.6$ \\\hline
DiffPool                & $75.27 \pm 0.8$ & $73.73 \pm 0.6$  &  $74.23 \pm 1.0$ \\\hline
SAGPool$_h$             & $69.41 \pm 1.1$ & $69.12 \pm 0.5$  &  $73.33 \pm 0.8$ \\\hline

\end{tabular}
}
\vspace{-3mm}
\label{tab:graph_classification}
\end{table}

\subsection{Performance}

The model performance on node classification is reported in Table~\ref{tab:node_classification}.
The results show that the GNN models provided by tf\_geometric can achieve competitive performance with the official implementations.
Particularly, although tf\_geometric adopts a Transformer-based GAT algorithm rather than the official version, it still obtains almost the same accuracy scores reported in~\cite{DBLP:conf/iclr/VelickovicCCRLB18}.

For the graph classification task, the results are listed in Table~\ref{tab:graph_classification}.
Since the architectures of some models are adjusted, the model performance is sometimes better than that reported in the literature.
Note that by optimizing the architecture, the naive Mean-Max Pool outperforms some other graph pooling models in some cases.

\section{Conclusions}

We introduce tf\_geometric, an efficient and friendly library for graph deep learning, which is compatible with both TensorFlow 1.x and 2.x.
tf\_geometric provides kernel libraries for building graph neural networks as well as implementations of popular GNNs.
In particular, the kernel libraries consist of infrastructures for building efficient GNNs, which enable tf\_geometric to support single-graph computation, multi-graph computation, graph mini-batch, distributed training, etc.
Therefore, tf\_geometric can be used for a variety of graph deep learning tasks, such as transductive node classification, inductive node classification, link prediction, and graph classification.
tf\_geometric exposes both OOP APIs and Functional APIs, with which users can deal with advanced graph deep learning tasks.
Moreover, the APIs are friendly, and they are suitable for both beginners and experts.
We are actively working to further optimize the kernel libraries and integrate more existing GNN models for tf\_geometric.
In the future, we will keep tf\_geometric up-to-date with the latest research findings of GNNs and continually integrate future models into it.


\bibliographystyle{ACM-Reference-Format}
\bibliography{TFG}


\begin{thebibliography}{35}


\ifx \showCODEN    \undefined \def \showCODEN     #1{\unskip}     \fi
\ifx \showDOI      \undefined \def \showDOI       #1{#1}\fi
\ifx \showISBNx    \undefined \def \showISBNx     #1{\unskip}     \fi
\ifx \showISBNxiii \undefined \def \showISBNxiii  #1{\unskip}     \fi
\ifx \showISSN     \undefined \def \showISSN      #1{\unskip}     \fi
\ifx \showLCCN     \undefined \def \showLCCN      #1{\unskip}     \fi
\ifx \shownote     \undefined \def \shownote      #1{#1}          \fi
\ifx \showarticletitle \undefined \def \showarticletitle #1{#1}   \fi
\ifx \showURL      \undefined \def \showURL       {\relax}        \fi
\providecommand\bibfield[2]{#2}
\providecommand\bibinfo[2]{#2}
\providecommand\natexlab[1]{#1}
\providecommand\showeprint[2][]{arXiv:#2}

\bibitem[\protect\citeauthoryear{Bastings, Titov, Aziz, Marcheggiani, and
  Sima'an}{Bastings et~al\mbox{.}}{2017}]%
        {DBLP:conf/emnlp/BastingsTAMS17}
\bibfield{author}{\bibinfo{person}{Jasmijn Bastings}, \bibinfo{person}{Ivan
  Titov}, \bibinfo{person}{Wilker Aziz}, \bibinfo{person}{Diego Marcheggiani},
  {and} \bibinfo{person}{Khalil Sima'an}.} \bibinfo{year}{2017}\natexlab{}.
\newblock \showarticletitle{Graph Convolutional Encoders for Syntax-aware
  Neural Machine Translation}. In \bibinfo{booktitle}{\emph{Proceedings of the
  2017 Conference on Empirical Methods in Natural Language Processing, {EMNLP}
  2017, Copenhagen, Denmark, September 9-11, 2017}}.
  \bibinfo{publisher}{Association for Computational Linguistics},
  \bibinfo{pages}{1957--1967}.
\newblock


\bibitem[\protect\citeauthoryear{Borgwardt, Ong, Sch{\"o}nauer, Vishwanathan,
  Smola, and Kriegel}{Borgwardt et~al\mbox{.}}{2005}]%
        {Borgwardt2005ProteinFP}
\bibfield{author}{\bibinfo{person}{K. Borgwardt}, \bibinfo{person}{Cheng~Soon
  Ong}, \bibinfo{person}{S. Sch{\"o}nauer}, \bibinfo{person}{S. Vishwanathan},
  \bibinfo{person}{Alex Smola}, {and} \bibinfo{person}{H. Kriegel}.}
  \bibinfo{year}{2005}\natexlab{}.
\newblock \showarticletitle{Protein function prediction via graph kernels}.
\newblock \bibinfo{journal}{\emph{Bioinformatics}}  \bibinfo{volume}{21 Suppl
  1} (\bibinfo{year}{2005}), \bibinfo{pages}{i47--56}.
\newblock


\bibitem[\protect\citeauthoryear{Cui, Henrickson, Ke, and Wang}{Cui
  et~al\mbox{.}}{2020}]%
        {DBLP:journals/tits/CuiHKW20}
\bibfield{author}{\bibinfo{person}{Zhiyong Cui}, \bibinfo{person}{Kristian
  Henrickson}, \bibinfo{person}{Ruimin Ke}, {and} \bibinfo{person}{Yinhai
  Wang}.} \bibinfo{year}{2020}\natexlab{}.
\newblock \showarticletitle{Traffic Graph Convolutional Recurrent Neural
  Network: {A} Deep Learning Framework for Network-Scale Traffic Learning and
  Forecasting}.
\newblock \bibinfo{journal}{\emph{{IEEE} Trans. Intell. Transp. Syst.}}
  \bibinfo{volume}{21}, \bibinfo{number}{11} (\bibinfo{year}{2020}),
  \bibinfo{pages}{4883--4894}.
\newblock


\bibitem[\protect\citeauthoryear{Dobson and Doig}{Dobson and Doig}{2003}]%
        {Dobson2003DistinguishingES}
\bibfield{author}{\bibinfo{person}{P. Dobson} {and} \bibinfo{person}{A. Doig}.}
  \bibinfo{year}{2003}\natexlab{}.
\newblock \showarticletitle{Distinguishing enzyme structures from non-enzymes
  without alignments.}
\newblock \bibinfo{journal}{\emph{Journal of molecular biology}}
  \bibinfo{volume}{330 4} (\bibinfo{year}{2003}), \bibinfo{pages}{771--83}.
\newblock


\bibitem[\protect\citeauthoryear{Fan, Ma, Li, He, Zhao, Tang, and Yin}{Fan
  et~al\mbox{.}}{2019}]%
        {DBLP:conf/www/Fan0LHZTY19}
\bibfield{author}{\bibinfo{person}{Wenqi Fan}, \bibinfo{person}{Yao Ma},
  \bibinfo{person}{Qing Li}, \bibinfo{person}{Yuan He},
  \bibinfo{person}{Yihong~Eric Zhao}, \bibinfo{person}{Jiliang Tang}, {and}
  \bibinfo{person}{Dawei Yin}.} \bibinfo{year}{2019}\natexlab{}.
\newblock \showarticletitle{Graph Neural Networks for Social Recommendation}.
  In \bibinfo{booktitle}{\emph{The World Wide Web Conference, {WWW} 2019, San
  Francisco, CA, USA, May 13-17, 2019}}. \bibinfo{publisher}{{ACM}},
  \bibinfo{pages}{417--426}.
\newblock


\bibitem[\protect\citeauthoryear{Fang, Sun, Gan, Pillai, Wang, and Liu}{Fang
  et~al\mbox{.}}{2020}]%
        {DBLP:conf/emnlp/FangSGPWL20}
\bibfield{author}{\bibinfo{person}{Yuwei Fang}, \bibinfo{person}{Siqi Sun},
  \bibinfo{person}{Zhe Gan}, \bibinfo{person}{Rohit Pillai},
  \bibinfo{person}{Shuohang Wang}, {and} \bibinfo{person}{Jingjing Liu}.}
  \bibinfo{year}{2020}\natexlab{}.
\newblock \showarticletitle{Hierarchical Graph Network for Multi-hop Question
  Answering}. In \bibinfo{booktitle}{\emph{Proceedings of the 2020 Conference
  on Empirical Methods in Natural Language Processing, {EMNLP} 2020, Online,
  November 16-20, 2020}}. \bibinfo{publisher}{Association for Computational
  Linguistics}, \bibinfo{pages}{8823--8838}.
\newblock


\bibitem[\protect\citeauthoryear{Fey and Lenssen}{Fey and Lenssen}{2019}]%
        {DBLP:journals/corr/abs-1903-02428}
\bibfield{author}{\bibinfo{person}{Matthias Fey} {and}
  \bibinfo{person}{Jan~Eric Lenssen}.} \bibinfo{year}{2019}\natexlab{}.
\newblock \showarticletitle{Fast Graph Representation Learning with PyTorch
  Geometric}.
\newblock \bibinfo{journal}{\emph{CoRR}}  \bibinfo{volume}{abs/1903.02428}
  (\bibinfo{year}{2019}).
\newblock
\urldef\tempurl%
\url{http://arxiv.org/abs/1903.02428}
\showURL{%
\tempurl}


\bibitem[\protect\citeauthoryear{Finn, Abbeel, and Levine}{Finn
  et~al\mbox{.}}{2017}]%
        {DBLP:conf/icml/FinnAL17}
\bibfield{author}{\bibinfo{person}{Chelsea Finn}, \bibinfo{person}{Pieter
  Abbeel}, {and} \bibinfo{person}{Sergey Levine}.}
  \bibinfo{year}{2017}\natexlab{}.
\newblock \showarticletitle{Model-Agnostic Meta-Learning for Fast Adaptation of
  Deep Networks}. In \bibinfo{booktitle}{\emph{Proceedings of the 34th
  International Conference on Machine Learning, {ICML} 2017, Sydney, NSW,
  Australia, 6-11 August 2017}} \emph{(\bibinfo{series}{Proceedings of Machine
  Learning Research}, Vol.~\bibinfo{volume}{70})}. \bibinfo{publisher}{{PMLR}},
  \bibinfo{pages}{1126--1135}.
\newblock


\bibitem[\protect\citeauthoryear{Fout, Byrd, Shariat, and Ben{-}Hur}{Fout
  et~al\mbox{.}}{2017}]%
        {DBLP:conf/nips/FoutBSB17}
\bibfield{author}{\bibinfo{person}{Alex Fout}, \bibinfo{person}{Jonathon Byrd},
  \bibinfo{person}{Basir Shariat}, {and} \bibinfo{person}{Asa Ben{-}Hur}.}
  \bibinfo{year}{2017}\natexlab{}.
\newblock \showarticletitle{Protein Interface Prediction using Graph
  Convolutional Networks}. In \bibinfo{booktitle}{\emph{Advances in Neural
  Information Processing Systems 30: Annual Conference on Neural Information
  Processing Systems 2017, December 4-9, 2017, Long Beach, CA, {USA}}}.
  \bibinfo{pages}{6530--6539}.
\newblock


\bibitem[\protect\citeauthoryear{Gaudelet, Day, Jamasb, Soman, Regep, Liu,
  Hayter, Vickers, Roberts, Tang, Roblin, Blundell, Bronstein, and
  Taylor{-}King}{Gaudelet et~al\mbox{.}}{2020}]%
        {DBLP:journals/corr/abs-2012-05716}
\bibfield{author}{\bibinfo{person}{Thomas Gaudelet}, \bibinfo{person}{Ben Day},
  \bibinfo{person}{Arian~R. Jamasb}, \bibinfo{person}{Jyothish Soman},
  \bibinfo{person}{Cristian Regep}, \bibinfo{person}{Gertrude Liu},
  \bibinfo{person}{Jeremy B.~R. Hayter}, \bibinfo{person}{Richard Vickers},
  \bibinfo{person}{Charles Roberts}, \bibinfo{person}{Jian Tang},
  \bibinfo{person}{David Roblin}, \bibinfo{person}{Tom~L. Blundell},
  \bibinfo{person}{Michael~M. Bronstein}, {and} \bibinfo{person}{Jake~P.
  Taylor{-}King}.} \bibinfo{year}{2020}\natexlab{}.
\newblock \showarticletitle{Utilising Graph Machine Learning within Drug
  Discovery and Development}.
\newblock \bibinfo{journal}{\emph{CoRR}}  \bibinfo{volume}{abs/2012.05716}
  (\bibinfo{year}{2020}).
\newblock
\urldef\tempurl%
\url{https://arxiv.org/abs/2012.05716}
\showURL{%
\tempurl}


\bibitem[\protect\citeauthoryear{Grattarola and Alippi}{Grattarola and
  Alippi}{2021}]%
        {DBLP:journals/cim/GrattarolaA21}
\bibfield{author}{\bibinfo{person}{Daniele Grattarola} {and}
  \bibinfo{person}{Cesare Alippi}.} \bibinfo{year}{2021}\natexlab{}.
\newblock \showarticletitle{Graph Neural Networks in TensorFlow and Keras with
  Spektral [Application Notes]}.
\newblock \bibinfo{journal}{\emph{{IEEE} Comput. Intell. Mag.}}
  \bibinfo{volume}{16}, \bibinfo{number}{1} (\bibinfo{year}{2021}),
  \bibinfo{pages}{99--106}.
\newblock


\bibitem[\protect\citeauthoryear{Hu, Qian, Fang, and Xu}{Hu
  et~al\mbox{.}}{2019}]%
        {DBLP:conf/mm/HuQFX19}
\bibfield{author}{\bibinfo{person}{Jun Hu}, \bibinfo{person}{Shengsheng Qian},
  \bibinfo{person}{Quan Fang}, {and} \bibinfo{person}{Changsheng Xu}.}
  \bibinfo{year}{2019}\natexlab{}.
\newblock \showarticletitle{Hierarchical Graph Semantic Pooling Network for
  Multi-modal Community Question Answer Matching}. In
  \bibinfo{booktitle}{\emph{Proceedings of the 27th {ACM} International
  Conference on Multimedia, {MM} 2019, Nice, France, October 21-25, 2019}}.
  \bibinfo{publisher}{{ACM}}, \bibinfo{pages}{1157--1165}.
\newblock


\bibitem[\protect\citeauthoryear{Kipf and Welling}{Kipf and Welling}{2017}]%
        {DBLP:conf/iclr/KipfW17}
\bibfield{author}{\bibinfo{person}{Thomas~N. Kipf} {and} \bibinfo{person}{Max
  Welling}.} \bibinfo{year}{2017}\natexlab{}.
\newblock \showarticletitle{Semi-Supervised Classification with Graph
  Convolutional Networks}. In \bibinfo{booktitle}{\emph{5th International
  Conference on Learning Representations, {ICLR} 2017, Toulon, France, April
  24-26, 2017, Conference Track Proceedings}}.
  \bibinfo{publisher}{OpenReview.net}.
\newblock


\bibitem[\protect\citeauthoryear{Klicpera, Bojchevski, and
  G{\"{u}}nnemann}{Klicpera et~al\mbox{.}}{2019}]%
        {DBLP:conf/iclr/KlicperaBG19}
\bibfield{author}{\bibinfo{person}{Johannes Klicpera},
  \bibinfo{person}{Aleksandar Bojchevski}, {and} \bibinfo{person}{Stephan
  G{\"{u}}nnemann}.} \bibinfo{year}{2019}\natexlab{}.
\newblock \showarticletitle{Predict then Propagate: Graph Neural Networks meet
  Personalized PageRank}. In \bibinfo{booktitle}{\emph{7th International
  Conference on Learning Representations, {ICLR} 2019, New Orleans, LA, USA,
  May 6-9, 2019}}. \bibinfo{publisher}{OpenReview.net}.
\newblock


\bibitem[\protect\citeauthoryear{Lee, Lee, and Kang}{Lee et~al\mbox{.}}{2019}]%
        {DBLP:conf/icml/LeeLK19}
\bibfield{author}{\bibinfo{person}{Junhyun Lee}, \bibinfo{person}{Inyeop Lee},
  {and} \bibinfo{person}{Jaewoo Kang}.} \bibinfo{year}{2019}\natexlab{}.
\newblock \showarticletitle{Self-Attention Graph Pooling}. In
  \bibinfo{booktitle}{\emph{Proceedings of the 36th International Conference on
  Machine Learning, {ICML} 2019, 9-15 June 2019, Long Beach, California,
  {USA}}} \emph{(\bibinfo{series}{Proceedings of Machine Learning Research},
  Vol.~\bibinfo{volume}{97})}. \bibinfo{publisher}{{PMLR}},
  \bibinfo{pages}{3734--3743}.
\newblock


\bibitem[\protect\citeauthoryear{Li, Gan, Cheng, and Liu}{Li
  et~al\mbox{.}}{2019}]%
        {DBLP:conf/iccv/LiGCL19}
\bibfield{author}{\bibinfo{person}{Linjie Li}, \bibinfo{person}{Zhe Gan},
  \bibinfo{person}{Yu Cheng}, {and} \bibinfo{person}{Jingjing Liu}.}
  \bibinfo{year}{2019}\natexlab{}.
\newblock \showarticletitle{Relation-Aware Graph Attention Network for Visual
  Question Answering}. In \bibinfo{booktitle}{\emph{2019 {IEEE/CVF}
  International Conference on Computer Vision, {ICCV} 2019, Seoul, Korea
  (South), October 27 - November 2, 2019}}. \bibinfo{publisher}{{IEEE}},
  \bibinfo{pages}{10312--10321}.
\newblock


\bibitem[\protect\citeauthoryear{Li, Qian, Zhang, and Liu}{Li
  et~al\mbox{.}}{2020}]%
        {DBLP:journals/bigdata/LiQZL20}
\bibfield{author}{\bibinfo{person}{Yang Li}, \bibinfo{person}{Buyue Qian},
  \bibinfo{person}{Xianli Zhang}, {and} \bibinfo{person}{Hui Liu}.}
  \bibinfo{year}{2020}\natexlab{}.
\newblock \showarticletitle{Graph Neural Network-Based Diagnosis Prediction}.
\newblock \bibinfo{journal}{\emph{Big Data}} \bibinfo{volume}{8},
  \bibinfo{number}{5} (\bibinfo{year}{2020}), \bibinfo{pages}{379--390}.
\newblock


\bibitem[\protect\citeauthoryear{Marcheggiani, Bastings, and
  Titov}{Marcheggiani et~al\mbox{.}}{2018}]%
        {DBLP:conf/naacl/MarcheggianiBT18}
\bibfield{author}{\bibinfo{person}{Diego Marcheggiani},
  \bibinfo{person}{Jasmijn Bastings}, {and} \bibinfo{person}{Ivan Titov}.}
  \bibinfo{year}{2018}\natexlab{}.
\newblock \showarticletitle{Exploiting Semantics in Neural Machine Translation
  with Graph Convolutional Networks}. In \bibinfo{booktitle}{\emph{Proceedings
  of the 2018 Conference of the North American Chapter of the Association for
  Computational Linguistics: Human Language Technologies, NAACL-HLT, New
  Orleans, Louisiana, USA, June 1-6, 2018, Volume 2 (Short Papers)}}.
  \bibinfo{publisher}{Association for Computational Linguistics},
  \bibinfo{pages}{486--492}.
\newblock


\bibitem[\protect\citeauthoryear{Pfaff, Fortunato, Sanchez-Gonzalez, and
  Battaglia}{Pfaff et~al\mbox{.}}{2021}]%
        {pfaff2021learning}
\bibfield{author}{\bibinfo{person}{Tobias Pfaff}, \bibinfo{person}{Meire
  Fortunato}, \bibinfo{person}{Alvaro Sanchez-Gonzalez}, {and}
  \bibinfo{person}{Peter Battaglia}.} \bibinfo{year}{2021}\natexlab{}.
\newblock \showarticletitle{Learning Mesh-Based Simulation with Graph
  Networks}. In \bibinfo{booktitle}{\emph{International Conference on Learning
  Representations}}.
\newblock
\urldef\tempurl%
\url{https://openreview.net/forum?id=roNqYL0_XP}
\showURL{%
\tempurl}


\bibitem[\protect\citeauthoryear{Ranjan, Sanyal, and Talukdar}{Ranjan
  et~al\mbox{.}}{2020}]%
        {DBLP:conf/aaai/RanjanST20}
\bibfield{author}{\bibinfo{person}{Ekagra Ranjan}, \bibinfo{person}{Soumya
  Sanyal}, {and} \bibinfo{person}{Partha~P. Talukdar}.}
  \bibinfo{year}{2020}\natexlab{}.
\newblock \showarticletitle{{ASAP:} Adaptive Structure Aware Pooling for
  Learning Hierarchical Graph Representations}. In
  \bibinfo{booktitle}{\emph{The Thirty-Fourth {AAAI} Conference on Artificial
  Intelligence, {AAAI} 2020, The Thirty-Second Innovative Applications of
  Artificial Intelligence Conference, {IAAI} 2020, The Tenth {AAAI} Symposium
  on Educational Advances in Artificial Intelligence, {EAAI} 2020, New York,
  NY, USA, February 7-12, 2020}}. \bibinfo{publisher}{{AAAI} Press},
  \bibinfo{pages}{5470--5477}.
\newblock


\bibitem[\protect\citeauthoryear{Sampathkumar}{Sampathkumar}{2021}]%
        {DBLP:journals/corr/abs-2101-02870}
\bibfield{author}{\bibinfo{person}{Vishnu~Ram Sampathkumar}.}
  \bibinfo{year}{2021}\natexlab{}.
\newblock \showarticletitle{ADiag: Graph Neural Network Based Diagnosis of
  Alzheimer's Disease}.
\newblock \bibinfo{journal}{\emph{CoRR}}  \bibinfo{volume}{abs/2101.02870}
  (\bibinfo{year}{2021}).
\newblock
\urldef\tempurl%
\url{https://arxiv.org/abs/2101.02870}
\showURL{%
\tempurl}


\bibitem[\protect\citeauthoryear{Sanchez{-}Gonzalez, Godwin, Pfaff, Ying,
  Leskovec, and Battaglia}{Sanchez{-}Gonzalez et~al\mbox{.}}{2020}]%
        {DBLP:conf/icml/Sanchez-Gonzalez20}
\bibfield{author}{\bibinfo{person}{Alvaro Sanchez{-}Gonzalez},
  \bibinfo{person}{Jonathan Godwin}, \bibinfo{person}{Tobias Pfaff},
  \bibinfo{person}{Rex Ying}, \bibinfo{person}{Jure Leskovec}, {and}
  \bibinfo{person}{Peter~W. Battaglia}.} \bibinfo{year}{2020}\natexlab{}.
\newblock \showarticletitle{Learning to Simulate Complex Physics with Graph
  Networks}. In \bibinfo{booktitle}{\emph{Proceedings of the 37th International
  Conference on Machine Learning, {ICML} 2020, 13-18 July 2020, Virtual Event}}
  \emph{(\bibinfo{series}{Proceedings of Machine Learning Research},
  Vol.~\bibinfo{volume}{119})}. \bibinfo{publisher}{{PMLR}},
  \bibinfo{pages}{8459--8468}.
\newblock


\bibitem[\protect\citeauthoryear{Sen, Namata, Bilgic, Getoor, Gallagher, and
  Eliassi{-}Rad}{Sen et~al\mbox{.}}{2008}]%
        {DBLP:journals/aim/SenNBGGE08}
\bibfield{author}{\bibinfo{person}{Prithviraj Sen}, \bibinfo{person}{Galileo
  Namata}, \bibinfo{person}{Mustafa Bilgic}, \bibinfo{person}{Lise Getoor},
  \bibinfo{person}{Brian Gallagher}, {and} \bibinfo{person}{Tina
  Eliassi{-}Rad}.} \bibinfo{year}{2008}\natexlab{}.
\newblock \showarticletitle{Collective Classification in Network Data}.
\newblock \bibinfo{journal}{\emph{{AI} Mag.}} \bibinfo{volume}{29},
  \bibinfo{number}{3} (\bibinfo{year}{2008}), \bibinfo{pages}{93--106}.
\newblock


\bibitem[\protect\citeauthoryear{Tan, Liu, Zhao, Yang, Zhou, and Hu}{Tan
  et~al\mbox{.}}{2020}]%
        {DBLP:conf/www/TanLZYZH20}
\bibfield{author}{\bibinfo{person}{Qiaoyu Tan}, \bibinfo{person}{Ninghao Liu},
  \bibinfo{person}{Xing Zhao}, \bibinfo{person}{Hongxia Yang},
  \bibinfo{person}{Jingren Zhou}, {and} \bibinfo{person}{Xia Hu}.}
  \bibinfo{year}{2020}\natexlab{}.
\newblock \showarticletitle{Learning to Hash with Graph Neural Networks for
  Recommender Systems}. In \bibinfo{booktitle}{\emph{{WWW} '20: The Web
  Conference 2020, Taipei, Taiwan, April 20-24, 2020}}.
  \bibinfo{publisher}{{ACM} / {IW3C2}}, \bibinfo{pages}{1988--1998}.
\newblock


\bibitem[\protect\citeauthoryear{Velickovic, Cucurull, Casanova, Romero,
  Li{\`{o}}, and Bengio}{Velickovic et~al\mbox{.}}{2018}]%
        {DBLP:conf/iclr/VelickovicCCRLB18}
\bibfield{author}{\bibinfo{person}{Petar Velickovic}, \bibinfo{person}{Guillem
  Cucurull}, \bibinfo{person}{Arantxa Casanova}, \bibinfo{person}{Adriana
  Romero}, \bibinfo{person}{Pietro Li{\`{o}}}, {and} \bibinfo{person}{Yoshua
  Bengio}.} \bibinfo{year}{2018}\natexlab{}.
\newblock \showarticletitle{Graph Attention Networks}. In
  \bibinfo{booktitle}{\emph{6th International Conference on Learning
  Representations, {ICLR} 2018, Vancouver, BC, Canada, April 30 - May 3, 2018,
  Conference Track Proceedings}}. \bibinfo{publisher}{OpenReview.net}.
\newblock


\bibitem[\protect\citeauthoryear{Velickovic, Fedus, Hamilton, Li{\`{o}},
  Bengio, and Hjelm}{Velickovic et~al\mbox{.}}{2019}]%
        {DBLP:conf/iclr/VelickovicFHLBH19}
\bibfield{author}{\bibinfo{person}{Petar Velickovic}, \bibinfo{person}{William
  Fedus}, \bibinfo{person}{William~L. Hamilton}, \bibinfo{person}{Pietro
  Li{\`{o}}}, \bibinfo{person}{Yoshua Bengio}, {and} \bibinfo{person}{R.~Devon
  Hjelm}.} \bibinfo{year}{2019}\natexlab{}.
\newblock \showarticletitle{Deep Graph Infomax}. In
  \bibinfo{booktitle}{\emph{7th International Conference on Learning
  Representations, {ICLR} 2019, New Orleans, LA, USA, May 6-9, 2019}}.
  \bibinfo{publisher}{OpenReview.net}.
\newblock


\bibitem[\protect\citeauthoryear{Vinyals, Bengio, and Kudlur}{Vinyals
  et~al\mbox{.}}{2016}]%
        {DBLP:journals/corr/VinyalsBK15}
\bibfield{author}{\bibinfo{person}{Oriol Vinyals}, \bibinfo{person}{Samy
  Bengio}, {and} \bibinfo{person}{Manjunath Kudlur}.}
  \bibinfo{year}{2016}\natexlab{}.
\newblock \showarticletitle{Order Matters: Sequence to sequence for sets}. In
  \bibinfo{booktitle}{\emph{4th International Conference on Learning
  Representations, {ICLR} 2016, San Juan, Puerto Rico, May 2-4, 2016,
  Conference Track Proceedings}}.
\newblock
\urldef\tempurl%
\url{http://arxiv.org/abs/1511.06391}
\showURL{%
\tempurl}


\bibitem[\protect\citeauthoryear{Wale, Watson, and Karypis}{Wale
  et~al\mbox{.}}{2008}]%
        {DBLP:journals/kais/WaleWK08}
\bibfield{author}{\bibinfo{person}{Nikil Wale}, \bibinfo{person}{Ian~A.
  Watson}, {and} \bibinfo{person}{George Karypis}.}
  \bibinfo{year}{2008}\natexlab{}.
\newblock \showarticletitle{Comparison of descriptor spaces for chemical
  compound retrieval and classification}.
\newblock \bibinfo{journal}{\emph{Knowl. Inf. Syst.}} \bibinfo{volume}{14},
  \bibinfo{number}{3} (\bibinfo{year}{2008}), \bibinfo{pages}{347--375}.
\newblock


\bibitem[\protect\citeauthoryear{Wang, Yu, Zheng, Gan, Gai, Ye, Li, Zhou,
  Huang, Ma, Huang, Guo, Zhang, Lin, Zhao, Li, Smola, and Zhang}{Wang
  et~al\mbox{.}}{2019}]%
        {DBLP:journals/corr/abs-1909-01315}
\bibfield{author}{\bibinfo{person}{Minjie Wang}, \bibinfo{person}{Lingfan Yu},
  \bibinfo{person}{Da Zheng}, \bibinfo{person}{Quan Gan}, \bibinfo{person}{Yu
  Gai}, \bibinfo{person}{Zihao Ye}, \bibinfo{person}{Mufei Li},
  \bibinfo{person}{Jinjing Zhou}, \bibinfo{person}{Qi Huang},
  \bibinfo{person}{Chao Ma}, \bibinfo{person}{Ziyue Huang},
  \bibinfo{person}{Qipeng Guo}, \bibinfo{person}{Hao Zhang},
  \bibinfo{person}{Haibin Lin}, \bibinfo{person}{Junbo Zhao},
  \bibinfo{person}{Jinyang Li}, \bibinfo{person}{Alexander~J. Smola}, {and}
  \bibinfo{person}{Zheng Zhang}.} \bibinfo{year}{2019}\natexlab{}.
\newblock \showarticletitle{Deep Graph Library: Towards Efficient and Scalable
  Deep Learning on Graphs}.
\newblock \bibinfo{journal}{\emph{CoRR}}  \bibinfo{volume}{abs/1909.01315}
  (\bibinfo{year}{2019}).
\newblock
\urldef\tempurl%
\url{http://arxiv.org/abs/1909.01315}
\showURL{%
\tempurl}


\bibitem[\protect\citeauthoryear{Wang, Zhang, Liu, Liu, Zhang, Lin, and
  Zha}{Wang et~al\mbox{.}}{2020b}]%
        {DBLP:conf/www/WangZLLZLZ20}
\bibfield{author}{\bibinfo{person}{Wen Wang}, \bibinfo{person}{Wei Zhang},
  \bibinfo{person}{Shukai Liu}, \bibinfo{person}{Qi Liu}, \bibinfo{person}{Bo
  Zhang}, \bibinfo{person}{Leyu Lin}, {and} \bibinfo{person}{Hongyuan Zha}.}
  \bibinfo{year}{2020}\natexlab{b}.
\newblock \showarticletitle{Beyond Clicks: Modeling Multi-Relational Item Graph
  for Session-Based Target Behavior Prediction}. In
  \bibinfo{booktitle}{\emph{{WWW} '20: The Web Conference 2020, Taipei, Taiwan,
  April 20-24, 2020}}. \bibinfo{publisher}{{ACM} / {IW3C2}},
  \bibinfo{pages}{3056--3062}.
\newblock


\bibitem[\protect\citeauthoryear{Wang, Ma, Wang, Jin, Wang, Tang, Jia, and
  Yu}{Wang et~al\mbox{.}}{2020a}]%
        {DBLP:conf/www/Wang0WJWTJY20}
\bibfield{author}{\bibinfo{person}{Xiaoyang Wang}, \bibinfo{person}{Yao Ma},
  \bibinfo{person}{Yiqi Wang}, \bibinfo{person}{Wei Jin}, \bibinfo{person}{Xin
  Wang}, \bibinfo{person}{Jiliang Tang}, \bibinfo{person}{Caiyan Jia}, {and}
  \bibinfo{person}{Jian Yu}.} \bibinfo{year}{2020}\natexlab{a}.
\newblock \showarticletitle{Traffic Flow Prediction via Spatial Temporal Graph
  Neural Network}. In \bibinfo{booktitle}{\emph{{WWW} '20: The Web Conference
  2020, Taipei, Taiwan, April 20-24, 2020}}. \bibinfo{publisher}{{ACM} /
  {IW3C2}}, \bibinfo{pages}{1082--1092}.
\newblock


\bibitem[\protect\citeauthoryear{Wu, Jr., Zhang, Fifty, Yu, and Weinberger}{Wu
  et~al\mbox{.}}{2019b}]%
        {DBLP:conf/icml/WuSZFYW19}
\bibfield{author}{\bibinfo{person}{Felix Wu}, \bibinfo{person}{Amauri H.~Souza
  Jr.}, \bibinfo{person}{Tianyi Zhang}, \bibinfo{person}{Christopher Fifty},
  \bibinfo{person}{Tao Yu}, {and} \bibinfo{person}{Kilian~Q. Weinberger}.}
  \bibinfo{year}{2019}\natexlab{b}.
\newblock \showarticletitle{Simplifying Graph Convolutional Networks}. In
  \bibinfo{booktitle}{\emph{Proceedings of the 36th International Conference on
  Machine Learning, {ICML} 2019, 9-15 June 2019, Long Beach, California,
  {USA}}} \emph{(\bibinfo{series}{Proceedings of Machine Learning Research},
  Vol.~\bibinfo{volume}{97})}. \bibinfo{publisher}{{PMLR}},
  \bibinfo{pages}{6861--6871}.
\newblock


\bibitem[\protect\citeauthoryear{Wu, He, and Xu}{Wu et~al\mbox{.}}{2019a}]%
        {DBLP:conf/kdd/WuHX19}
\bibfield{author}{\bibinfo{person}{Jun Wu}, \bibinfo{person}{Jingrui He}, {and}
  \bibinfo{person}{Jiejun Xu}.} \bibinfo{year}{2019}\natexlab{a}.
\newblock \showarticletitle{DEMO-Net: Degree-specific Graph Neural Networks for
  Node and Graph Classification}. In \bibinfo{booktitle}{\emph{Proceedings of
  the 25th {ACM} {SIGKDD} International Conference on Knowledge Discovery {\&}
  Data Mining, {KDD} 2019, Anchorage, AK, USA, August 4-8, 2019}}.
  \bibinfo{publisher}{{ACM}}, \bibinfo{pages}{406--415}.
\newblock


\bibitem[\protect\citeauthoryear{Ying, You, Morris, Ren, Hamilton, and
  Leskovec}{Ying et~al\mbox{.}}{2018}]%
        {DBLP:conf/nips/YingY0RHL18}
\bibfield{author}{\bibinfo{person}{Zhitao Ying}, \bibinfo{person}{Jiaxuan You},
  \bibinfo{person}{Christopher Morris}, \bibinfo{person}{Xiang Ren},
  \bibinfo{person}{William~L. Hamilton}, {and} \bibinfo{person}{Jure
  Leskovec}.} \bibinfo{year}{2018}\natexlab{}.
\newblock \showarticletitle{Hierarchical Graph Representation Learning with
  Differentiable Pooling}. In \bibinfo{booktitle}{\emph{Advances in Neural
  Information Processing Systems 31: Annual Conference on Neural Information
  Processing Systems 2018, NeurIPS 2018, December 3-8, 2018, Montr{\'{e}}al,
  Canada}}. \bibinfo{pages}{4805--4815}.
\newblock


\bibitem[\protect\citeauthoryear{Zhang, Cui, Neumann, and Chen}{Zhang
  et~al\mbox{.}}{2018}]%
        {DBLP:conf/aaai/ZhangCNC18}
\bibfield{author}{\bibinfo{person}{Muhan Zhang}, \bibinfo{person}{Zhicheng
  Cui}, \bibinfo{person}{Marion Neumann}, {and} \bibinfo{person}{Yixin Chen}.}
  \bibinfo{year}{2018}\natexlab{}.
\newblock \showarticletitle{An End-to-End Deep Learning Architecture for Graph
  Classification}. In \bibinfo{booktitle}{\emph{Proceedings of the
  Thirty-Second {AAAI} Conference on Artificial Intelligence, (AAAI-18), the
  30th innovative Applications of Artificial Intelligence (IAAI-18), and the
  8th {AAAI} Symposium on Educational Advances in Artificial Intelligence
  (EAAI-18), New Orleans, Louisiana, USA, February 2-7, 2018}}.
  \bibinfo{publisher}{{AAAI} Press}, \bibinfo{pages}{4438--4445}.
\newblock


\end{thebibliography}










\typeout{get arXiv to do 4 passes: Label(s) may have changed. Rerun}
\end{document}